\documentclass[10pt,twocolumn,letterpaper]{article}
\usepackage[pagenumbers]{cvpr}
\usepackage{iftex}
\RequirePDFTeX

\usepackage{cuted} 

\usepackage{currfile} 

\usepackage{caption} 

\usepackage{amsmath,amssymb,amsthm}
\usepackage{booktabs,multirow}
\usepackage{xspace}
\newtheorem{proposition}{Proposition}
\newtheorem{corollary}{Corollary}
\newtheorem{observation}{Observation}
\newcommand{\xh}{\hat{x}}
\newcommand{\zh}{\hat{z}_0}
\newcommand{\Sd}{S_{d}}
\newcommand{\Su}{S_{u}}
\newcommand{\sm}{\mathrm{softplus}}

\definecolor{cvprblue}{rgb}{0.21,0.49,0.74}
\usepackage{xr-hyper}
\usepackage[breaklinks,colorlinks,allcolors=cvprblue]{hyperref}
\hypersetup{
  pdftitle={SurvAD: How Anomalies Survive Diffusion Reconstruction},
  pdfauthor={Lexing Zhang},
}
\usepackage{microtype}

\def\confName{CVPR}
\def\confYear{2027}

\title{SurvAD: How Anomalies Survive Diffusion Reconstruction}

\author{Lexing Zhang\\
Shanghai University}

\begin{document}
\maketitle
\begin{abstract}
Diffusion-based anomaly detectors reconstruct a test image towards normality, yet how much of an anomaly survives, its \emph{leak}, remains uncharacterized.
We show that a first-order deterministic sampler obeys an exact recursion along the anomaly, whose continuous limit is a \emph{survival law}.
The anomaly decays exponentially as the noise falls, at a rate set by the \emph{retention} $r$, the share the model keeps on the states it visits.
Because $r$ depends on the noise level and not on the step count, one measured path predicts the leak of other step counts, samplers and schedules from a nearby start to within a few hundredths, on four diffusion models and real defects.
The law also shows that the leak's growth over steps is exposure bias and that normal details obey the same law, so changing the path or the training trades leak for detail error.
We introduce \emph{SurvAD}, which fine-tunes the reconstructor on inference-like starts and fits a comparator that adds evidence to the residual map without suppressing its ranking elsewhere.
For under $2\%$ more inference time, SurvAD raises the multi-class pixel average precision (P-AP) of DeCo-Diff from $76.8$ to $80.7$ on MVTec AD and from $57.5$ to $60.0$ on VisA.
It matches CostFilter-AD on the same reconstructor with $9\times$ fewer trainable parameters and, under Dinomaly's protocol, exceeds its reported P-AP on Dinomaly on both benchmarks.
\end{abstract}
\section{Introduction}
\label{sec:intro}

Industrial visual inspection needs detectors that are trained on normal images alone and localize defects at test time~\cite{bergmann2019mvtec}.
Reconstruction-based anomaly detection repairs a test image towards normality and flags where the repair differs from the input~\cite{zavrtanik2021draem}.
Diffusion models are natural repairers, since they denoise a partially noised or corrupted version of the input~\cite{wyatt2022anoddpm,mousakhan2023ddad,he2024diad,yao2024glad}, and the strongest of them, DeCo-Diff~\cite{beizaee2025deco}, reaches a multi-class pixel average precision (P-AP) of $76.8$ on MVTec AD.
Yet every diffusion detector faces the same dilemma: a faithful reconstruction keeps a scratch on wood, a loose one also smooths the grain around it, and both errors appear in the residual.
Keeping the anomaly is the \emph{leak}.

\begin{figure}[t]
\centering
\includegraphics[width=\linewidth]{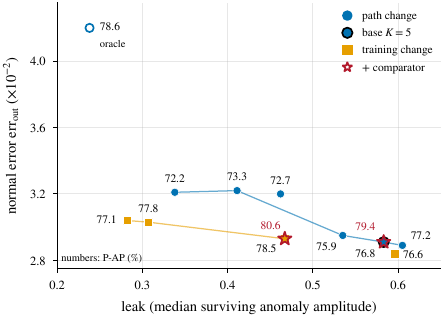}
\caption{\textbf{Changing the reconstructor trades leak for normal error; SurvAD's comparator raises P-AP off this curve.} On MVTec AD with DeCo-Diff, path changes move along one curve and training changes shift it; numbers are P-AP. The hollow point is the oracle, regenerating inside the ground-truth mask. The comparator (stars) leaves leak and normal error unchanged and raises P-AP from $76.8$ to $79.4$ on the base and from $78.5$ to $80.6$ with start matching (seed 0).}
\label{fig:curve}
\end{figure}

Existing detectors address this dilemma in three ways, and each leaves it open.
One family shortens the sampling path or adapts it per image~\cite{song2021ddim,yao2024glad}, but a shorter path leaks less only by reconstructing normal detail less accurately (\cref{fig:curve}).
A second family retrains or conditions the denoiser~\cite{zhang2023diffusionad,yao2024glad}, which lowers the leak on the states it visits, although at test time those states still differ from the ones seen in training~\cite{ning2023input,ning2024elucidating}.
A third family, led by CostFilter-AD~\cite{zhang2025costfilter}, filters a cost volume between the input and its reconstructions, but the filtered map alone falls below the residual in most settings and needs a hand-set blend with it (\cref{sec:cf}).
No law says how far the first two can trade one error for the other, or how a learned score should treat the residual.
We derive that law and build a detector, \emph{SurvAD}, from it.

Along the anomaly, each step of a first-order deterministic sampler mixes what the state still carries with what the prediction keeps, in a ratio fixed by the noise schedule.
In the continuous limit the anomaly therefore decays exponentially as the noise falls, at a rate set by the \emph{retention}, the share the model keeps on the states it visits.
We show that on the sampler's own states the retention depends on the noise level and not on the number of steps, so measured on one path it predicts the leak of other step counts, samplers and schedules from a nearby start to within $0.025$, on four diffusion models.

This law has three consequences, the first being that the multi-step growth of the leak is exposure bias~\cite{ning2023input,ning2024elucidating}.
Second, the same law governs normal details, so every change of sampling path or training trades leak for detail error, and across six path and four training changes none of these reconstructors exceeds $78.7$ P-AP (\cref{fig:curve}).
A fixed residual must therefore blur.
Third, a partly repaired defect looks like detail error, so a learned scorer must add evidence to the residual rather than replace it, and we fit it on normals the reconstructor has not memorized, whose detail error resembles that of test normals.

\begin{figure*}[t]
\centering
\includegraphics[width=\linewidth]{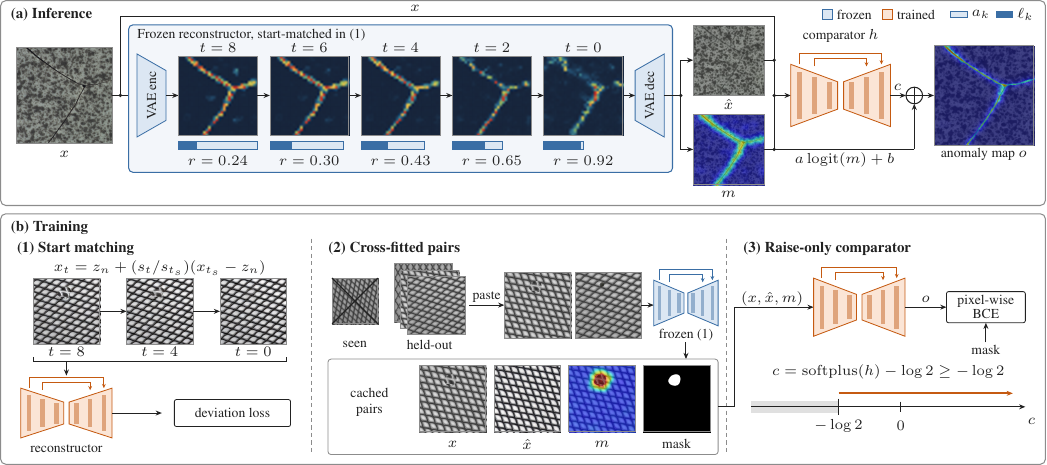}
\caption{\textbf{SurvAD keeps the sampling path and adds evidence to its residual.}
(a) At test time a start-matched reconstructor maps $x$ to $\xh$ and the residual $m$; thumbnails show the latent deviation per step, and bars the amplitude $a_k$, the kept part $\ell_k$ and the retention $r$ (medians, seed 0).
A raise-only comparator reads $(x,\xh,m)$ and raises the residual logit where it finds evidence.
(b) Offline, start matching fine-tunes the reconstructor on inference-like starts, and the comparator trains on cross-fitted pairs, synthetic anomalies on held-out normals passed through the frozen reconstructor.}
\label{fig:method}
\end{figure*}

We apply these rules in SurvAD, which keeps the sampling path and changes what the detector does around it (\cref{fig:method}).
We fine-tune the reconstructor on inference-like starts, which lowers the leak by about a fifth, then freeze it and reconstruct synthetic anomalies on normals it has not seen, forming cross-fitted pairs that expose the normal part of the residual.
On these pairs we train a raise-only comparator, which reads the input, reconstruction and residual and adds a correction with a fixed floor to the residual's logit, so it can move a pixel up the ranking but cannot suppress one except by raising others.
Unlike CostFilter-AD, which filters costs between DINO features and needs a hand-set blend weight, our comparator acts on the residual's logit with no blend weight and about $9\times$ fewer trainable parameters.
We add no second sampling pass and under $2\%$ to the inference time, yet we raise DeCo-Diff's P-AP to $80.7$ on MVTec AD and from $57.5$ to $60.0$ on VisA, where the score holds when the reconstructor is retrained without the comparator's normals.
SurvAD matches CostFilter-AD applied to DeCo-Diff in the mean, and under the protocol of Dinomaly~\cite{guo2025dinomaly} it exceeds the P-AP that CostFilter-AD reports on Dinomaly on both benchmarks.

Our contributions are as follows.
\begin{itemize}
\item The anomaly survival law, which predicts the leak of unseen paths (\cref{sec:law}).
\item Its dual for normal details, which shows why a fixed residual must blur (\cref{sec:dual}).
\item The add-evidence rule for learned scores, and a measured memorization gap that favors held-out normals for pixel comparators (\cref{sec:memo,sec:cf}).
\item \emph{SurvAD}, a detector built from start matching and a cross-fitted raise-only comparator (\cref{sec:start,sec:crossfit,sec:comparator}).
\end{itemize}

\section{Related Work}
\label{sec:related}

\paragraph{Diffusion reconstruction for anomaly detection.}
AnoDDPM~\cite{wyatt2022anoddpm} and DDAD~\cite{mousakhan2023ddad} reconstruct from a partially noised input.
DiffusionAD~\cite{zhang2023diffusionad} predicts the clean image in one step and adds a discriminative segmentation head trained with synthetic anomalies.
DiAD~\cite{he2024diad} conditions a latent diffusion model on the input through a semantic-guided network.
GLAD~\cite{yao2024glad} trains the denoiser on noised anomalous inputs and selects the sampling step per image.
DeCo-Diff~\cite{beizaee2025deco} predicts the deviation from normality and samples from the clean latent of the test image.
Its deviation sampler has the same form as the DDIM update~\cite{song2021ddim}, a first-order integration of the probability-flow ODE~\cite{song2021score}.
GLAD uses DDIM directly, and guidance only changes the denoiser it integrates~\cite{dhariwal2021diffusion}.

\paragraph{Feature reconstruction.}
In the multi-class setting one model serves every class, which removes the cost of one model per class but makes the normal distribution harder to cover~\cite{beizaee2025deco,guo2025dinomaly}.
Dinomaly~\cite{guo2025dinomaly} reconstructs the features of a frozen encoder, and its noisy bottleneck and loose reconstruction keep the decoder from copying its input.

\paragraph{Exposure bias.}
A diffusion sampler visits states produced by its own predictions, which differ from the forward-process states it was trained on~\cite{ning2023input}.
Input perturbation~\cite{ning2023input} and epsilon scaling~\cite{ning2024elucidating} reduce the gap for generation.
In anomaly detection the gap is directional: what matters is the anomaly component of the state.
Training on the states the sampler visits is the DAgger argument~\cite{ross2011dagger}.

\paragraph{Inversion and editing.}
Reconstruction-based detection is a selective edit: keep every normal detail and remove the anomaly.
SDEdit~\cite{meng2022sdedit} trades realism for faithfulness through the start noise.
Exact inversion~\cite{mokady2023null,wallace2023edict} maximizes faithfulness, so it keeps the anomaly.
InvAD~\cite{sakai2026invad} skips reconstruction, inverts the input, and scores its deviation from the normal latent prior.

\paragraph{Learned comparison.}
DRAEM~\cite{zavrtanik2021draem} and its descendants learn a segmentation network on pairs of a synthetic anomaly and its reconstruction.
PNI~\cite{bae2023pni} refines anomaly maps with a network trained on synthetic anomalies to follow image edges, and OneNIP~\cite{gao2024onenip} trains a supervised refiner on real normal and synthesized anomalous images.
CostFilter-AD~\cite{zhang2025costfilter} filters a matching-cost volume between the input and its reconstructions or normal templates, an idea from cost-volume filtering~\cite{hosni2013fast}, guided by input features and an initial anomaly map.
It treats the anomaly map as a noisy matching signal and filters it.
CostFilter-AD, like DRAEM, trains its head on images the reconstructor has seen.

\paragraph{Memory banks and cross-fitting.}
Memory-bank detectors score a test patch by its distance to stored normal patches~\cite{roth2022patchcore,defard2021padim}, and BoundarySupport~\cite{chae2026boundary} finds that clean patches miss the near-defect regions a test image contains.
Cross-fitting, \ie fitting a nuisance model on data held out from the main model, is standard in semiparametric estimation~\cite{chernozhukov2018double}, but to our knowledge it has not been used for scorers on a reconstructor.
\section{The Survival Law and SurvAD}
\label{sec:method}

\emph{SurvAD} maps a test image to a pixel anomaly map with a frozen diffusion reconstructor, whose sampling path it leaves as it is (\cref{fig:method}).
Let $x$ be a test image, $\xh$ its reconstruction and $m\in[0,1)^{H\times W}$ the residual map.
At test time both come from a start-matched reconstructor, and a comparator $h$ reads $(x,\xh,m)$ and scores each pixel.
Training has three parts.
Start matching fine-tunes the reconstructor on inference-like starts, which lowers the exposure bias that inflates the leak.
The frozen reconstructor then reconstructs cross-fitted pairs, synthetic anomalies on normals it has not seen, so the comparator can learn the normal part of the residual.
The raise-only comparator is trained on these pairs to add evidence to the residual's logit, keeping the residual's ranking where it finds none.
The comparator applies to any frozen reconstructor with a pixel output, and start matching to any diffusion reconstructor with a deterministic sampler.

\subsection{Setting}
\label{sec:setting}
A latent diffusion reconstructor encodes the test image to $z$, builds a start state $x_{t_0}$ from it, and runs $K$ deterministic steps $t_0 > t_1 > \dots > t_{K-1}$.
At each visited state $x_k$ the model predicts a clean latent $\zh^{(k)} = f(x_k, t_k)$, and the sampler moves to
\begin{equation}
\begin{aligned}
x_{k+1} &= \sqrt{\bar\alpha_{k+1}}\,\zh^{(k)} + \sqrt{1-\bar\alpha_{k+1}}\,\hat\epsilon_k,\\
\hat\epsilon_k &= \frac{x_k-\sqrt{\bar\alpha_k}\,\zh^{(k)}}{\sqrt{1-\bar\alpha_k}},
\end{aligned}
\label{eq:ddim}
\end{equation}
(DDIM~\cite{song2021ddim}).
The output is $\xh = D(\zh^{(K-1)})$.
The base reconstructor~\cite{beizaee2025deco} uses a UNet with attention as its backbone and operates in the latent space of a frozen VAE.
It samples with its own deviation sampler $x_{k+1} = \zh^{(k)} + \tfrac{s_{k+1}}{s_k}(x_k - \zh^{(k)})$, which has the same form.
Let $z = z_n + d$ with $d$ supported on the anomaly, and measure every quantity along $d$ with the linear projection $P(v) = \langle v, d\rangle / \langle d, d\rangle$ on the anomaly region.
Since this needs $d$, amplitudes are measured on $395$ synthetic anomalies pasted on MVTec AD test normals.
Write $\sigma_t = \sqrt{1-\bar\alpha_t}/\sqrt{\bar\alpha_t}$ and $\lambda = 1/\sigma$ ($\sqrt{\mathrm{SNR}}$).
At step $k$ the state carries the amplitude $a_k = P(x_k/\sqrt{\bar\alpha_k} - z_n)$ and the prediction keeps $\ell_k = P(\zh^{(k)} - z_n)$.
The output leak is $L_K = \ell_{K-1}$, and the \emph{retention} $r_k = \ell_k / a_k$ is the share of the anomaly the model keeps on the state it actually visits.
The anomaly map is a residual between $\xh$ and the input, so it sees the anomaly with strength about $|d|(1-\Sd)$, where the survival $\Sd = L_K/a_0$ is the share of the anomaly kept.

\subsection{An exact recursion}
\label{sec:recursion}
\begin{proposition}[Amplitude recursion]
\label{prop:recursion}
For any denoiser $f$, any input and any noise,
\begin{equation}
\begin{aligned}
a_{k+1} &= \rho_k\,a_k + (1-\rho_k)\,\ell_k,\\
\rho_k &= \sigma_{t_{k+1}}/\sigma_{t_k}\in(0,1).
\end{aligned}
\label{eq:recursion}
\end{equation}
For that sampler the same holds with $\rho_k = s_{k+1}/s_k$. With fresh noise in the step, as in DDIM with $\eta>0$, it holds with a smaller $\rho_k$ plus a zero-mean term.
\end{proposition}
\begin{proof}
Divide \cref{eq:ddim} by $\sqrt{\bar\alpha_{k+1}}$, substitute $\hat\epsilon_k$: $x_{k+1}/\sqrt{\bar\alpha_{k+1}} = \zh^{(k)} + \rho_k\,(x_k/\sqrt{\bar\alpha_k} - \zh^{(k)})$. Subtract $z_n$ and apply $P$.
\end{proof}
The identity is exact for any denoiser; the empirical content is in the retention $r_k$.
A perfect repair at step $k$, where $\ell_k = 0$, still leaves $\rho_k a_k$ in the next state: the sampler re-injects the anomaly through $\hat\epsilon_k$.
Unrolling with $r_k$ gives
\begin{equation}
L_K = r_{K-1}\prod_{j<K-1}\big[\rho_j + (1-\rho_j)\,r_j\big].
\label{eq:unroll}
\end{equation}
\begin{corollary}[Leak growth implies rising retention]
\label{cor:growth}
If $r_j \le r_0 < 1$ for all $j$, then $L_K < L_1$ for every $K\ge 2$. Hence $L_K > L_1$ implies a rising retention.
\end{corollary}
\begin{proof}
Each factor of \cref{eq:unroll} is $<1$, and $r_{K-1}\le r_0 = L_1$.
\end{proof}
On the base model the leak does grow, from $L_1 = 0.338$ to $L_5 = 0.583$ (\cref{tab:law}), and the retention on its own states rises from $0.34$ to $0.95$.
Feeding the oracle retention through \cref{eq:unroll} instead predicts $0.390$: exposure bias accounts for $0.193$ of the base leak and for $79\%$ of the growth from $L_1$ to $L_5$.
The multi-step growth of the leak thus comes from states the model was not trained on, \ie from exposure bias~\cite{ning2023input}.
Retention depends on the noise level and on the kind of state, visited or trained on, not on the step count that led there: nine- and five-step paths agree to within $0.006$ at every noise level.

\subsection{The survival law}
\label{sec:law}
\begin{table}[t]
\centering\small
\caption{\textbf{The law measured and tested.} All entries are medians. Base rows: 395 synthetic anomalies on MVTec AD test normals, the same masks filled with a normal detail, and 1220 real defects. GLAD rows: the 395 synthetic anomalies. Top: retention on its own states and on the oracle trajectory with $K{=}5$, and of the normal detail, taken from the closest unseen normal. Middle: leak, predicted by \cref{eq:unroll} from the same inputs' $K{=}5$ retention alone, with interpolation in $\ln\lambda$, and measured. Real defects against the closest unseen normal. Bottom: GLAD, $K$-step paths from $t{=}650$, predicted from the $K{=}25$ retention and measured.}
\label{tab:law}
\setlength{\tabcolsep}{2.4pt}
\begin{tabular}{@{}llccccc@{}}
\toprule
Base & $t$ & 8 & 6 & 4 & 2 & 0\\
\midrule
& $r_k$ own & 0.338 & 0.445 & 0.599 & 0.783 & 0.954\\
& $r_k$ oracle traj. & 0.338 & 0.427 & 0.517 & 0.568 & 0.783\\
& $r_k$ own, detail & 0.839 & 0.883 & 0.927 & 0.971 & 0.991\\
\midrule
Path transfer & $K$ & 1 & 2 & 3 & 5 & 9\\
\midrule
& synth., predicted & 0.338 & 0.429 & 0.539 & 0.587 & 0.603\\
& synth., measured & 0.338 & 0.411 & 0.535 & 0.583 & 0.605\\
& oracle traj., pred. & -- & -- & -- & 0.390 & --\\
& real, predicted & 0.592 & 0.653 & 0.751 & 0.791 & 0.805\\
& real, measured & 0.592 & 0.642 & 0.756 & 0.790 & 0.803\\
\midrule
GLAD & $K$ & 1 & 2 & 5 & 10 & 25\\
\midrule
& synth., predicted & 0.200 & 0.202 & 0.215 & 0.222 & 0.231\\
& synth., measured & 0.203 & 0.209 & 0.220 & 0.221 & 0.227\\
\bottomrule
\end{tabular}
\end{table}
\begin{proposition}[Anomaly survival law]
\label{prop:survival}
In the continuous limit, \cref{eq:ddim} becomes the probability-flow ODE $dy/d\sigma = (y - \zh)/\sigma$ with $y = x/\sqrt{\bar\alpha}$~\cite{song2021score}. The visibility $\gamma = \lambda a$ satisfies $d\gamma/d\lambda = \ell$, and
\begin{equation}
a(\lambda_{\mathrm{end}}) = a(\lambda_0)\,\exp\!\Big(-\!\int_{\ln\lambda_0}^{\ln\lambda_{\mathrm{end}}}\big(1-r(\lambda)\big)\,d\ln\lambda\Big).
\label{eq:survival}
\end{equation}
\end{proposition}
\begin{proof}
Project the ODE on $d$: $da/d\sigma = (a-\ell)/\sigma$. With $\lambda = 1/\sigma$ this is $d(\lambda a)/d\lambda = \ell = r a$, i.e.\ $d\ln a/d\ln\lambda = -(1-r)$.
\end{proof}
Two observations follow.
\textbf{(i) Conservation.} Wherever the prediction does not over-repair ($\ell\ge0$), the visibility never decreases: a deterministic sampler cannot forget an anomaly, the property that makes deterministic inversion faithful~\cite{mokady2023null,wallace2023edict} and diffusion detection leak.
\textbf{(ii) The low-noise end removes little.} Every unit of $\ln\lambda$ counts equally. The model's last interval spans $31\%$ of the range, as much as its first, yet it keeps $0.95$ of the anomaly in that interval against $0.34$ in the first.
\paragraph{A test: path transfer.}
On the sampler's own states, $r$ depends on the noise level and not on the step count, whereas on the oracle trajectory it differs by exposure bias, so the retention curve of one step count predicts the leak of any other.
From the five-step curve alone, interpolated in $\ln\lambda$, \cref{eq:unroll} predicts $0.429$, $0.539$ and $0.603$ for two, three and nine steps; the model measures $0.411$, $0.535$ and $0.605$ (\cref{tab:law}); the median error per sample is at most $0.010$.
Real defects, compared against the closest unseen normal, pass the same test to within $0.011$.
It also predicts the direction of the effect: a coarse step applies the high-noise retention over its whole interval, so fewer steps leak less.
GLAD~\cite{yao2024glad} passes the same test to within $0.014$ with its plain sampler, DDIM with classifier-free guidance. The two pixel-space models also pass it. DDAD~\cite{mousakhan2023ddad}, trained per class on MVTec AD, passes to within $0.008$, and ADM~\cite{dhariwal2021diffusion}, trained on ImageNet only, passes to within $0.023$.
\paragraph{Held-out paths and samplers.}
The same curves predict held-out schedules, samplers, start levels and GLAD's fusion step to within $0.025$ (supplementary Sec.~\ref{S-sec:supp-heldout}); only clean starts below $t{=}7$ are over-predicted, by $0.03$ to $0.09$.

\begin{figure}[b]
\centering
\includegraphics[width=\linewidth]{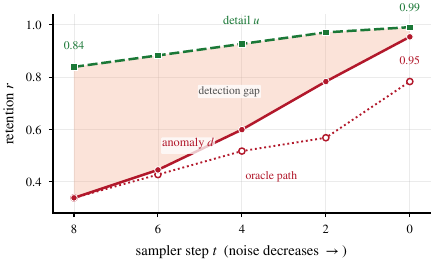}
\caption{\textbf{The dual survival law.} Retention $r$ of the anomaly and of a normal detail on the model's own states, for the base on MVTec AD. The detail is retained more at every noise level, and the shaded gap, not the leak alone, separates the two. The oracle path is where the model would sit without exposure bias, which start matching reduces.}
\label{fig:dual}
\end{figure}

\subsection{The dual survival law}
\label{sec:dual}
Nothing in \cref{prop:survival} uses the fact that $d$ is an anomaly.
For any direction $v$, the survival $S_v = a_v(\lambda_{\mathrm{end}})/a_v(\lambda_0)$ obeys \cref{eq:survival} with the retention $r_v$ of that direction.
Along the anomaly $d$ we want $\Sd\to 0$, since $\Sd$ is the leak.
Along a normal detail $u$, \eg the specific realization of a texture that cannot be inferred from context under noise, we want $\Su\to 1$.
A residual sees the anomaly with strength about $|d|(1-\Sd)$ and the detail with strength about $|u|(1-\Su)$, so detection depends on the gap between the two retention curves along the whole path, not on the leak alone (\cref{fig:dual}).

\begin{observation}[One curve]
\label{obs:curve}
At high noise both retentions are small: the model can infer neither the texture nor the anomaly and forgets both.
At low noise both approach one: the model is near the identity, keeping the detail and, through exposure bias, the anomaly.
A change of the sampling path, such as start time, number of steps, re-noising or masked regeneration, changes the interval of integration. Where it leaves the retention at a given noise level unchanged, $\Sd$ and $\Su$ move together.
For all classes, taking the aligned region of the closest unseen normal as a detail that the context does not determine, $r_u$ on the sampler's own states is also step-count invariant: the $K{=}5$ curve predicts $\Su$ for other step counts to within $0.003$. It exceeds $r_d$ at every noise level (\cref{tab:law}), and $\Su$ and $\Sd$ grow together with $K$; grid is the exception.
\end{observation}

This observation yields three predictions, all of which hold on the base model (\cref{fig:curve}).
\textbf{(a)} One step instead of five lowers both survivals. Detail-rich classes lose, carpet by $21.8$ and leather by $22.0$ P-AP, while leak-dominated ones gain, grid by $3.1$. Choosing $K$ per class on the ground truth reaches only $77.4$.
\textbf{(b)} Regenerating from high noise inside the ground-truth mask lowers $\Sd$ there and $\Su$ on the normal pixels covered by the dilated mask: transistor gains $20.1$, bottle loses $9.7$, and the mean gains only $1.8$.
\textbf{(c)} With a self-estimated mask, regeneration loses, reaching $72.7$ at best, because the mask misses the leaked defects.

\paragraph{Training shifts the curve.}
Training can change the retentions themselves, but not one without the other.
Training the denoiser on inference-like starts, which is start matching, lowers the leak from $0.583$ to $0.454\pm0.011$ with almost unchanged normal error and raises P-AP from $76.8$ to $78.6\pm0.2$.
Training it on its own visited states, the on-policy variant, lowers the leak further but raises the normal error. Accuracy falls to $77.8$ on all cells and $77.1$ on anomaly cells, with the largest losses on carpet and wood, $12.9$ and $8.7$ points. At low noise the network has a single deviation-removal function, and on stochastic textures the anomaly and the detail are locally indistinguishable to it.
Training moves the whole curve without separating the two survivals: across six path changes and four training changes, none of these reconstructors exceeds $78.7$ P-AP (supplementary Sec.~\ref{S-sec:supp-curve}), so the score is what remains.
To suppress the detail error, a fixed residual must average over space, and the averaging floods the normal pixels around each defect: when the false-positive rate on the other normal pixels is $1\%$ per class, $48\%$ of the normal pixels within $8$ pixels of a real defect are flagged.

\subsection{The memorization gap}
\label{sec:memo}
A learned comparator can use evidence the residual lacks, but to learn the normal part of the residual, $|u|(1-\Su)$, it must see normal reconstructions; the usual choice is reconstructions of the training images~\cite{zavrtanik2021draem,zhang2023diffusionad,zhang2025costfilter}.
On those images the reconstructor has memorized the details as well, so $\Su\approx 1$ and the detail error is invisible.
The base sets aside 362 of its MVTec AD training normals as a validation split. On these, the 99.9th percentile of the anomaly map averages $8.4$ on seen normals, $22.3$ on held-out normals and $24.4$ on test normals. Held-out normals are closer to test normals than seen ones in 14 of 15 classes.
For the feature reconstructor Dinomaly~\cite{guo2025dinomaly}, the same percentile is $5.4$ on seen and $7.5$ on test normals.
A comparator that reads the pixel residual of a diffusion reconstructor should therefore be fit on reconstructions of normals the reconstructor did not see, which is cross-fitting~\cite{chernozhukov2018double}; \cref{sec:cf} tests this against a refiner that compares DINO features.

\subsection{Start matching}
\label{sec:start}
A detection sampler starts from the test image, not from noise: the base reconstructor starts from the clean test latent at $t_s = 8$, while it trains its deviation predictor on heavily corrupted latents.
Its first states are thus unlike anything seen in training, and the retention on them rises above the oracle's.
We fine-tune the reconstructor, keeping its loss, and add a deviation loss on the states a perfect predictor would visit from an inference-like start: a clean latent with a synthetic anomaly pasted on whole cells of the latent grid, followed along the oracle trajectory $z_n + (s_k/s_{t_s})(x_{t_s}-z_n)$.
The result is off-policy training on matched starts, and the fine-tuned reconstructor is frozen for the next two parts.

\subsection{Cross-fitted pairs}
\label{sec:crossfit}
The comparator learns the normal part of the residual on normals the reconstructor has not memorized.
A normal image receives a synthetic anomaly, and the \emph{frozen} reconstructor runs its test-time path on it with the same start and number of steps, so the comparator sees the kind of $(\xh,m)$ it will score.
Each normal contributes a clean replica, its synthetic anomalies and a pixel mask.
The normals come from the reconstructor's held-out split: for the base on MVTec AD, the training normals it set aside.
Where a reconstructor holds out no normals, as for the base on VisA, GLAD and Dinomaly, we use normals it has seen; \cref{sec:ablation} compares the two.

\paragraph{Synthetic anomalies.}
We use Perlin-masked pastes of DTD textures~\cite{cimpoi2014dtd} or of the image itself, restricted to the object foreground, as in DRAEM~\cite{zavrtanik2021draem}.
Half of the anomalies are small ellipses.

\subsection{Raise-only comparator}
\label{sec:comparator}
The residual alone cannot separate a leak from a detail error, so the comparator reads the input for evidence the residual lacks.
For the base, $m$ is its geometric combination of image and latent residuals, squashed monotonically to $[0,1)$.
The comparator is a small U-Net $h$ with $4.9$\,M parameters applied to the concatenation of $x$, $\xh$ and $m$, with its last layer initialized to zero.
Its output logit is
\begin{equation}
o = a\,\mathrm{logit}(m) + b + \underbrace{\sm\big(h(x,\xh,m)\big) - \log 2}_{c\,\ge\,-\log 2},
\label{eq:raise}
\end{equation}
with scalars $a$ and $b$ learned jointly, initialized to $a=1$, $b=0$. The learned $a$ remains above one, so the recalibration is increasing.
At initialization $h = 0$, so $c = 0$ and $o$ is a monotone recalibration of the residual.
Training can lower $c$ by at most $\log 2$ but can raise it without bound: relative to that floor, the correction can only raise it.
When the comparator adds no evidence, so that $h\to-\infty$ and $c = -\log 2$, $o$ preserves the order of $m$: the comparator can move a pixel up the ranking but cannot push one down except by raising others.
At test time $o$ is the pixel's anomaly score.

\paragraph{Why a floor.}
A subtle defect that the reconstructor partly repaired leaves a residual resembling detail error, and synthetic anomalies cannot teach the difference; an unconstrained correction, $c = h$, therefore learns to erase such defects (\cref{sec:ablation}).

\paragraph{Loss.}
We train the comparator with pixel-wise binary cross-entropy on the clean and anomalous pairs.
\section{Experiments}
\label{sec:exp}
We first compare SurvAD with published detectors on MVTec AD and VisA, then validate the add-evidence rule and the comparator on a published refiner, ablate each design choice against the rule it tests, and finally trace where the gain comes from and what it costs.

\begin{table*}[t]
\centering\small
\caption{\textbf{Multi-class anomaly detection and localization.} I-AUROC / P-AUROC / P-AP / AUPRO (\%). $^\dagger$from DeCo-Diff~\cite{beizaee2025deco}. $^\ddagger$its reported run. --: not reported. Bold: best within each block. $\pm$: std over three runs. $^\S$Comparator on seen normals; strict cross-fitting: $59.3\pm0.2$ without and $60.3\pm0.2$ with start matching.}
\label{tab:main}
\setlength{\tabcolsep}{4pt}
\begin{tabular}{@{}lcccccccc@{}}
\toprule
& \multicolumn{4}{c}{MVTec AD} & \multicolumn{4}{c}{VisA}\\
\cmidrule(lr){2-5}\cmidrule(l){6-9}
Method & I-AUROC & P-AUROC & P-AP & AUPRO & I-AUROC & P-AUROC & P-AP & AUPRO\\
\midrule
RD4AD~\cite{deng2022rd4ad} & 94.6 & 96.1 & 48.6 & 91.1 & 92.4 & 98.1 & 38.0 & 91.8\\
UniAD~\cite{you2022uniad} & 96.5 & 96.8 & 43.4 & 90.7 & 88.8 & 98.3 & 33.7 & 85.5\\
SimpleNet~\cite{liu2023simplenet} & 95.3 & 96.9 & 45.9 & 86.5 & 87.2 & 96.8 & 34.7 & 81.4\\
DeSTSeg~\cite{zhang2023destseg} & 89.2 & 93.1 & 54.3 & 64.8 & 88.9 & 96.1 & 39.6 & 67.4\\
DiAD~\cite{he2024diad} & 97.2 & 96.8 & 52.6 & 90.7 & 86.8 & 96.0 & 26.1 & 75.2\\
ReContrast~\cite{guo2023recontrast} & 98.3 & 97.1 & 60.2 & 93.2 & 95.5 & 98.5 & 47.9 & 91.9\\
ViTAD~\cite{zhang2023vitad} & 98.3 & 97.7 & 55.3 & 91.4 & 90.5 & 98.2 & 36.6 & 85.1\\
MambaAD~\cite{he2024mambaad} & 98.6 & 97.7 & 56.3 & 93.1 & 94.3 & 98.5 & 39.4 & 91.0\\
OneNIP~\cite{gao2024onenip} & 97.9 & 97.9 & 63.7 & -- & 92.5 & 98.7 & 43.3 & --\\
GLAD~\cite{yao2024glad}$^\dagger$ & 97.5 & 97.4 & 60.8 & 93.0 & 91.8 & 97.4 & 34.6 & 92.0\\
Dinomaly~\cite{guo2025dinomaly} & 99.6 & 98.4 & 69.3 & 94.8 & 98.7 & 98.7 & 53.2 & 94.5\\
INP-Former~\cite{luo2025inpformer} & \textbf{99.7} & 98.5 & 71.0 & 94.9 & \textbf{98.9} & \textbf{98.9} & 51.2 & 94.4\\
Dinomaly + CostFilter-AD~\cite{zhang2025costfilter} & \textbf{99.7} & \textbf{98.7} & \textbf{75.6} & \textbf{95.6} & 98.8 & \textbf{98.9} & \textbf{59.9} & \textbf{94.7}\\
SwinAD~\cite{ninh2026swinad} & 99.4 & 98.1 & 74.4 & -- & 98.0 & 97.0 & 58.0 & --\\
DeCo-Diff~\cite{beizaee2025deco}$^\ddagger$ & 99.3 & 98.4 & 74.9 & 94.9 & 96.4 & 98.5 & 51.3 & 92.1\\
\midrule
DeCo-Diff, $K{=}5$ & 99.2 & 98.8 & 76.8 & 95.5 & 96.1 & 98.3 & 57.5 & 92.2\\
\quad SurvAD w/o comparator & 99.0 & 98.9 & 78.6{\scriptsize$\pm$0.2} & 95.8 & 96.1 & \textbf{98.7} & 57.3{\scriptsize$\pm$0.3} & \textbf{93.7}\\
\quad SurvAD w/o start matching & \textbf{99.4} & 98.9 & 79.2{\scriptsize$\pm$0.2} & 96.0 & 96.1 & 98.4 & \textbf{60.2}{\scriptsize$\pm$0.3}$^\S$ & 92.3\\
\midrule
SurvAD (ours) & 99.2 & \textbf{99.0} & \textbf{80.7}{\scriptsize$\pm$0.1} & \textbf{96.2} & 96.1 & \textbf{98.7} & 60.0{\scriptsize$\pm$0.4}$^\S$ & 93.5\\
\bottomrule
\end{tabular}
\end{table*}

\paragraph{Protocol.}
We use MVTec AD~\cite{bergmann2019mvtec} (15 classes) and VisA~\cite{zou2022visa} (12 classes) in the multi-class (unified) setting, with one model for all classes.
All variants keep the base's input view and its geometric anomaly map.
We report image AUROC (I-AUROC, map maximum), pixel AUROC (P-AUROC), pixel average precision (P-AP), and AUPRO up to a false-positive rate of $0.3$.
P-AP is average precision over all test pixels of a class, averaged unweighted over classes; AUPRO is averaged the same way.
P-AP is the primary localization metric because defect pixels are rare.
Baseline numbers are those of the cited papers, with RD4AD to MambaAD taken from Dinomaly's re-evaluation~\cite{guo2025dinomaly}.
The reconstructor is the released DeCo-Diff checkpoint with five steps ($K{=}5$); training details are in supplementary Sec.~\ref{S-sec:supp-impl}.
The released VisA checkpoint holds out no normals, so its comparators use normals it has seen; \cref{sec:ablation} retrains it without them.

\subsection{Main Results}
\begin{figure}[t]
\centering
\includegraphics[width=\linewidth]{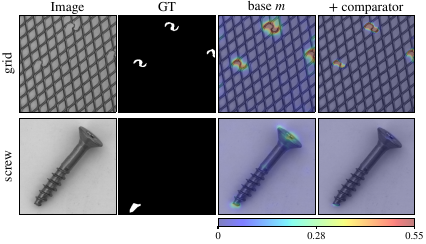}
\caption{\textbf{Qualitative maps of SurvAD's comparator} on DeCo-Diff. From left to right: the image, the ground-truth mask, the base residual and the comparator. The comparator keeps the residual's detections and narrows them to the defect: on grid it suppresses spurious nodes, on screw it removes the response at the head and keeps the tip.}
\label{fig:qual}
\end{figure}

SurvAD raises every pixel-level metric of DeCo-Diff on both benchmarks (\cref{tab:main}), and its two parts compose: on MVTec AD, start matching alone raises P-AP from $76.8$ to $78.6\pm0.2$, the comparator alone to $79.2\pm0.2$, and both to $80.7\pm0.1$, the highest P-AP in the table.
As a whole detector under Dinomaly's protocol, SurvAD scores $80.6\pm0.1$ and $61.1\pm0.4$, against the $75.6$ and $59.9$ that CostFilter-AD reports on Dinomaly.
The comparator on DeCo-Diff improves 13 of 15 classes, most on thin defects in textures and regular structures: grid $+8.2$, wood $+5.0$, tile $+3.5$, hazelnut $+3.5$ (\cref{fig:qual}).
On VisA the comparator adds $2.7$ P-AP to each reconstructor ($57.5\to60.2\pm0.3$ and $57.3\pm0.3\to60.0\pm0.4$) and improves 9 of 12 classes on each.
With the comparator, start matching adds $1.5$ P-AP on MVTec AD and $1.0\pm0.2$ on the VisA reconstructor retrained for strict cross-fitting (\cref{sec:ablation}); on the released VisA checkpoint it changes P-AP by $-0.2$ with and without the comparator.
Start matching dominates the one-time cost, $3.0$--$3.7$\,h on MVTec AD and $1.8$--$1.9$\,h on VisA, against about $0.8$\,h for the pair cache and comparator.

\subsection{Validation on a Published Refiner}
\label{sec:cf}
CostFilter-AD~\cite{zhang2025costfilter}, the strongest published refiner, follows the add-evidence rule only through a hand-set blend: it outputs $\lambda f + (1-\lambda)\,m$, where $m$ is the base map, $f\in[0,1]$ its filtered map, and $\lambda = 0.05$ on MVTec AD and $0.1$ on VisA and BTAD~\cite{mishra2021btad}.
Since $f\ge0$, this lowers $m$ by at most the factor $1-\lambda$, adds evidence on top, and preserves the order of $m$ wherever $f=0$: a manually set analogue of \cref{eq:raise}, with the bound in probability instead of logit.
The rule predicts that $f$ alone, which may also lower $m$, misses partly repaired defects; it falls below $m$ in four of the five settings.
Applied to the base, $f$ alone falls below the residual on both benchmarks ($73.9$ and $48.9$), while the blend gains $3.7$ and $1.5$.
Since both refiners only add evidence, they compose: blending $f$ into SurvAD's comparator map gives $81.3$ and $60.7$.
With its usual training on seen normals, the blended CostFilter-AD reaches $80.5$ and $59.0$, against $80.7$ and $60.0$ for SurvAD.
Paired by seed, the differences ($+0.19\pm0.61$ and $+0.95\pm1.64$) have $95\%$ intervals that contain zero, so the two match.
On the shared base the comparator alone scores $79.2$ and $60.2$; retrained on held-out normals over three seeds, the blend falls to $79.4\pm0.3$ on MVTec AD (paired $-1.1\pm0.4$), while the comparator gains $0.4\pm0.3$ from them.
The two effects differ by $1.5\pm0.4$ ($95\%$ interval $[0.6, 2.4]$), and on held-out normals the two refiners match (comparator minus blend $-0.2\pm0.3$, $95\%$ interval $[-1.0, 0.6]$).
The comparator has $4.9$\,M trainable parameters; CostFilter-AD's published plug-in adds $43.6$\,M, so it is about $9\times$ smaller.
On DeCo-Diff it also loads a frozen $85.8$\,M DINO encoder.

\begin{table}[t]
\centering\small
\caption{\textbf{Comparator ablations.} P-AP on MVTec AD at seed 0, with DeCo-Diff as the reconstructor.}
\label{tab:ablation}
\setlength{\tabcolsep}{2.5pt}
\begin{tabular}{@{}lcc@{}}
\toprule
Configuration & P-AP & $\Delta$\\
\midrule
full comparator & \textbf{79.4} & --\\
\midrule
correction unconstrained ($c = h$) & 75.2 & $-4.2$\\
\quad negative part only & 75.4 & $-4.0$\\
\quad positive part only & 78.8 & $-0.6$\\
seen training normals & 78.7 & $-0.7$\\
no small anomalies & 78.8 & $-0.6$\\
output smoothed & 77.9 & $-1.5$\\
\midrule
plain head, no residual & 69.4 & $-10.0$\\
\bottomrule
\end{tabular}
\end{table}

\subsection{Ablation Study}
\label{sec:ablation}
\paragraph{Raise-only correction and the residual.}
Removing the floor turns the comparator's gain into a loss: an unconstrained correction ($c=h$) scores $75.2$, below the residual's $76.8$; its negative part alone scores $75.4$, its positive part alone $78.8$, within $0.6$ of the full comparator.
A plain head without the residual scores $69.4$: the comparator re-ranks the residual rather than detecting from scratch.

\paragraph{Held-out normals.}
Training on seen instead of held-out normals costs $0.4\pm0.3$ P-AP over three seeds, seed 0 of \cref{tab:ablation} among them: the sign is the same at every seed, but the $95\%$ interval contains zero.
Normalizing the residual per pixel by held-out statistics fails ($61.2$).
On VisA, which has no held-out split, we retrain the reconstructor without the comparator's $864$ normals: there held-out normals add $0.25\pm0.24$ over seen ones, and the comparator scores $59.3\pm0.2$ on a base that the smaller training set lowers by $0.6$.
Start-matched on the other $7{,}795$ normals, the full detector scores $60.3\pm0.2$, and $61.1\pm0.2$ under Dinomaly's protocol, so its VisA scores hold when the comparator uses no normal the reconstructor has seen (supplementary Sec.~\ref{S-sec:supp-visa-strict}).

\paragraph{Anomaly size and output smoothing.}
Removing the small ellipses costs $0.6$ P-AP.
The head's map is sharper than the residual; smoothing it costs $1.5$ P-AP, and no non-learned alternative gains more than $0.1$.

\subsection{Further Analysis}
\label{sec:gain}
\paragraph{Source of the gain.}
We call the normal pixels within $8$ pixels of a defect its \emph{halo}.
Most of the comparator's gain comes from lifting defect cores above the blurred ring that the residual leaves around them, the ring predicted by the dual law (\cref{sec:dual}): with the halo excluded from the ranking, its gain on DeCo-Diff at seed 0 is $+0.5$ P-AP ($88.9\to89.4$) instead of $+2.6$, and AUPRO, which weighs each defect equally, rises from $95.5$ to $96.0$.
The rest comes from leaked defect pixels, whose reconstruction error is below the normal threshold; at a $1\%$ false-positive rate, the fraction of these pixels detected rises from $75.1\%$ to $77.4\%$.
Against CostFilter-AD on the same reconstructor, SurvAD's comparator flags fewer of these halo pixels at the same false-positive rate on both benchmarks ($-2.7$ and $-3.3$ points, $95\%$ intervals excluding zero), while CostFilter-AD finds more small and medium defects.

\paragraph{Inference cost.} The comparator adds $1.9\%$ to the inference time: at batch $16$ on one RTX 4090D the five-step detector takes $68.4$\,ms per image ($14.6$ images/s, $7.0$\,GB peak), $1.3$\,ms of it in the comparator. On one RTX 3090 at batch $16$, SurvAD takes $105.3$\,ms per image, $103.7$ of them in DeCo-Diff, against $15.4$\,ms for Dinomaly with CostFilter-AD: about $7\times$ the time for $+5.0$ and $+1.2$ P-AP under that refiner's protocol (supplementary Tab.~\ref{S-tab:cost-feat} and Fig.~\ref{S-fig:pareto}).
\section{Conclusion}
\label{sec:conclusion}
How much of an anomaly survives a first-order deterministic sampler is set by the retention on its own states, a function of the noise level rather than the step count. The same law governs normal details, so a fixed residual must blur.
Two rules follow: train on inference-like states, and let a learned scorer add to the residual map rather than replace it.
Built on these rules, \emph{SurvAD} matches the strongest published refiner applied to its reconstructor in the mean, with about $9\times$ fewer trainable parameters.

\paragraph{Limitations.}
The law predicts the leak from a measured retention, not the retention itself.
It is tested on MVTec AD against a proxy normal, and fails when the start is moved below $t{=}7$.
SurvAD gains in pixel ranking: its I-AUROC stays at DeCo-Diff's, $0.5$ and $2.7$ below Dinomaly with CostFilter-AD.
The gain from held-out normals has a $95\%$ interval that contains zero on both benchmarks, and on VisA strict cross-fitting needs a retrained reconstructor, on which start matching adds $1.0$ P-AP against $-0.2$ on the released one.
The comparator inherits the synthetic-anomaly training set and requires a pixel reconstruction; with Dinomaly it loses $1.8$ P-AP on VisA.
{
    \small
    \bibliographystyle{ieeenat_fullname}
    \bibliography{main}
}
\end{document}